\documentclass[runningheads]{llncs}
\usepackage{multirow} 
\usepackage{hyperref}
\usepackage[T1]{fontenc}
\usepackage{scrextend}
\usepackage{graphicx}
\usepackage{pgfplots}
\usepackage{comment}
\usepackage{svg}
\usepackage{amsmath,amssymb,amsfonts}
\usepackage{colortbl}
\usepackage{tabulary}

\begin{document}
\title{Influence of color correction on pathology detection in Capsule Endoscopy}
%
%
\author{Bidossessi Emmanuel Agossou \and
Marius Pedersen \and
Kiran Raja \and
Anuja Vats \and
Pål Anders Floor
}
%
%
\institute{Department of Computer Science, Norwegian University of Science and Technology (NTNU) 
\maketitle
\vspace{-1.5 em}

\begin{abstract}
Pathology detection in Wireless Capsule Endoscopy (WCE) using deep learning has been explored in the recent past. However, deep learning models can be influenced by the color quality of the dataset used to train them, impacting detection, segmentation and classification tasks. In this work, we evaluate the impact of color correction on pathology detection using two prominent object detection models: Retinanet and YOLOv5. We first generate two color corrected versions of a popular WCE dataset 
(i.e., SEE-AI dataset) using two different color correction functions. We then evaluate the performance of the Retinanet and YOLOv5 on the original and color corrected versions of the dataset. The results reveal that color correction makes the models generate larger bounding boxes and larger intersection areas with the ground truth annotations. Furthermore, color correction leads to an increased number of false positives for certain pathologies. However, these effects do not translate into a consistent improvement in performance metrics such as F1-scores, IoU, and AP50. The code is available at \href{https://github.com/agossouema2011/WCE2024}{https://github.com/agossouema2011/WCE2024}.

\end{abstract}
\section{Introduction}
Digestive system diseases are widespread around the world and cause considerable distress which can be fatal. In 2019, about 2276.27 million estimated prevalent cases and 2.56 million deaths were counted for digestive system diseases~\cite{wang2023global}. Endoscopy is the procedure commonly used to visualize the gastrointestinal tract and detect diseases \cite{bchir2019multiple}. However, traditional endoscopy is uncomfortable and painful for patients \cite{neilson2020patient}, discouraging it for wider preventive screening programs. A new type of gastrointestinal endoscopy, known as Wireless Capsule Endoscopy (WCE) \cite{iddan2000wireless}, was introduced in 2000, which is an alternative and minimally invasive screening method. In WCE, the patient swallows a pillcam (pill-sized camera) which moves through the gastrointestinal tract and records videos \cite{muruganantham2021survey}. The whole process can take on average eight to twelve hours, generating around 60,000 frames \cite{rahim2020survey}. Considering the long procedure duration, the demand placed on gastroenterologists for WCE screening is huge, leaving room for missing pathologies. Therefore, methods to improve pathology detection accuracy and reduce the time spent in screening WCE images are of great importance.

However, WCE has some drawbacks compared to traditional endoscopy like low frame rate and images of low resolution with poor quality \cite{sushma2022recent}. In addition, colors are distorted in WCE, making it challenging to perform color reproduction and color consistency \cite{watine2023enhancement}. Correcting or enhancing the color of the WCE images before training a deep learning model could help improve pathology detection accuracy. Faithfulness to color of human tissues and blood vessels is key to  accurately detecting abnormalities that often deviate from normal in their appearances \cite{watine2023enhancement}. Post-processing algorithms can distort colors that may lead to misinterpretation.  Thus making color reproduction accuracy and consistency all the more significant for human inspection as well as automated diagnosis. Color accuracy refers to the ability of a system to produce exact color matches from input to output, while color consistency refers to the ability to produce image data with a similar response  as that in a human interpreter \cite{badano2015consistency}\cite{watine2023enhancement}. Maintaining color accuracy through color correction along with resolution and contrast are important for  diagnostic use \cite{badano2015consistency}.

This work evaluates the influence of color correction on pathology detection in WCE. Detection in our context means pathology localization inside the image using bounding boxes. To the best of our knowledge,  little attention is given to this direction in the literature. The contributions of this work are the following:
\begin{itemize}
    \item We present two new color corrected versions of the publicly available SEE-AI \cite{yokote2024small} dataset. The dataset will be made publicly available to other researchers for investigating approaches for classification and detection tasks.
    \item We present a detailed analysis evaluating how slight changes in colors could potentially impact the behaviour of detection models in WCE. The analysis points to strengths and weakness of color correction for WCE for future works to build upon.
\end{itemize}

\section{Related works}
Object detection has gained huge attention \cite{liu2021unbiased} and consists of identifying objects or instances of objects in an image or a video using bounding boxes. There are two main groups of object detection methods: \textit{one-stage detectors and two-stage detectors}\cite{zhou2021instant}\cite{mi2022active}. One-stage objects detectors train models using a large amount of ground truth data. i.e., accurate human-annotated data. They require only a single pass through the neural network and predict all the bounding boxes in one go. One-stage methods include You Only Look Once (YOLO) \cite{redmon2016you}, and Single Shot Detector (SSD) \cite{liu2016ssd}.
The two-stage detectors on the other hand generate a set of regions of interest with a region proposal network, and then perform classification and bounding box regression. Popular two-stage models include Faster-RCNN \cite{ren2015faster} and Mask R-CNN~\cite{he2017mask}.

Ding et al.~\cite{ding2019gastroenterologist} used a Convolutional Neural Network (CNN) based  model to first classify WCE images into normal and abnormal frames, and further categorize the abnormal images into different pathology classes including inflammation, ulcer, polyps, lymphangiectasia, and bleeding.  Aoki et al. \cite{aoki2020automatic} used a CNN-based model to detect blood content in WCE images.
A CNN-based detection system for various abnormalities using WCE videos from multiple centers was developed by Aoki et al. \cite{aoki2021automatic}. The system combines four CNN-based models: three SSD models and one ResNet50, to detect among other, mucosal breaks, angioectasia, polyps, submucosal tumors, venous structures, blood contents.
Yokote et al.\cite{yokote2024small} created an annotated dataset for WCE containing 12 types of small bowel pathologies. In addition, an object detection model based on YOLOv5 was used to detect the pathologies.

Cherepkova  and  Hardeberg \cite{cherepkova2018enhancing} investigated the impact of image color enhancement on the performance of a lesion classification algorithm for skin cancer detection. Six different image enhancement algorithms were selected to investigate which algorithms have more influence on lesion detection improvement, focusing especially on color correction. A CNN based classifier and support vector machine were used on a set of dermoscopy image dataset.  Color enhancement generally yielded slightly better results for color versions of the images in their analysis \cite{cherepkova2018enhancing}. In order to solve color distortion problem in WCE, Watine et al. \cite{watine2023enhancement} considered the color correction of endoscopy videos through post-processing using a dedicated color checker to select the colors that typically appear in the human colon. The developed color checker based on the colon gamut was named ColonColorChecker (CCC). Two color correction matrices were derived, one based on a standard Gretag Macbeth ColorChecker (CC) and the other bases on the CCC. These were applied to enhance color in WCE images. From their objective and subjective evaluations with two experienced gastroenterologists, the color corrections provide a clear improvement of color in clinical WCE videos. However, the color corrections decreased the contrast in the color corrected videos reducing the diagnostic value of the approach \cite{watine2023enhancement}.

\section{Methodology}
Motivated by the limited attention to localizing and detecting the pathology when color correction is applied using CC and CCC, we present a series of analysis in this work. We make use of the publicly available WCE SEE-AI dataset~\cite{yokote2024small} and create two color corrected versions using CC and CCC \cite{watine2023enhancement}. As a result, we get two color corrected datasets. We then benchmark Retinanet and YOLOv5 models \cite{lin2017focal} on all the three datasets: the original SEE-AI, and the two color corrected, to understand the impact on pathology detection. We briefly present the WCE SEE-AI dataset \cite{yokote2024small} and relevant details of color correction. Further, we present our experimental set up and discuss relevant metrics used for analysis.

\subsection{Dataset} \textbf{SEE-AI Dataset:}
The SEE-AI dataset \cite{yokote2024small} is an annotated dataset for WCE with bounding box coordinates that contains 12 classes of small bowel pathologies. The images were collected with WCE PillCam SB-3 \cite{PillCamSB3}. The dataset contains 18,481 images extracted from 523 small bowel WCE procedures performed at Kyushu University Hospital (Japan). 12,320 images were annotated with 23,033 disease lesions, and combined with 6,161 normal images.
The SEE-AI dataset exhibits imbalance in pathology categories with fewer images of some pathologies (diverticulum, stenosis, SMT) as opposed to others (lymph-follicle, erosion, polyp-like)  as shown in Fig.~\ref{fig1}.

\textbf{Color Corrected Datasets:} Making use of two color correction matrices  proposed by \cite{watine2023enhancement}, i.e., \textit{ CC } and the \textit{ CCC} corrections, we create two additional datasets for our analysis. The CC was generated from the standard 24 patch color checker, while the CCC was generated from a selection of 30 colors (24 colors that appear within the gastrointestinal system and 6 gray scale). 
\begin{itemize}
    \item SEE-AI ColorChecker Corrected Dataset (CCD), which is the SEE-AI Dataset from \cite{yokote2024small} corrected with the CC. We refer to this dataset as CCD. 
    \item SEE-AI Colon Color Corrected Dataset (CCCD), which is the SEE-AI Dataset from \cite{yokote2024small} corrected with the CCC. We refer to this dataset as CCCD.
\end{itemize}
Fig.~\ref{fig2}  presents two samples of the original SEE-AI images (in the middle), and the corresponding CC images (on the left side) and CCC images (on the right side). As seen in  Fig.~\ref{fig2}, color correction leads to a significant enhancement in the colors, particularly the reddish hues. However, compared to the original images, a loss of contrast can be observed.

\subsection{Experimental Protocols}
Challenges faced by object detection methods with images include \textit{class imbalance} \cite{zhao2020rethinking,lin2017focal} and \textit{multiple aspect ratios} \cite{hsu2020ratio}. Class imbalance is when the dataset has only few labeled images for some classes, or when the spatial area of object of interest is smaller compared to the background. Multiple aspect ratios (aspect ratio imbalance) are when objects vary in aspect ratio with a variety of sizes: small, medium, and large objects with varying orientations\cite{analyticsindiamagAspectRatio}. These challenges make it difficult for object detection methods to achieve good detection results. Considering the nature of the dataset used for this analysis, we make use of Retinanet model as it is reported to have better performance under class imbalance problems due to incorporated focal loss \cite{lin2017focal}. To verify the consistency of our results, we also benchmark YOLOv5 with focal loss on our datasets.  We further study each of the two models in three different settings that focus on pathology detection:
\begin{enumerate}
    \item Retinanet and YOLOv5 on SEE-AI Dataset (R-OrigD)
    \item Retinanet and YOLOv5 on the CCD (R-CCD)
    \item Retinanet and YOLOv5 on the CCCD (R-CCCD)
    
\end{enumerate}

\begin{figure}[!htbp]
\centerline{
\includegraphics[width=1\textwidth]
{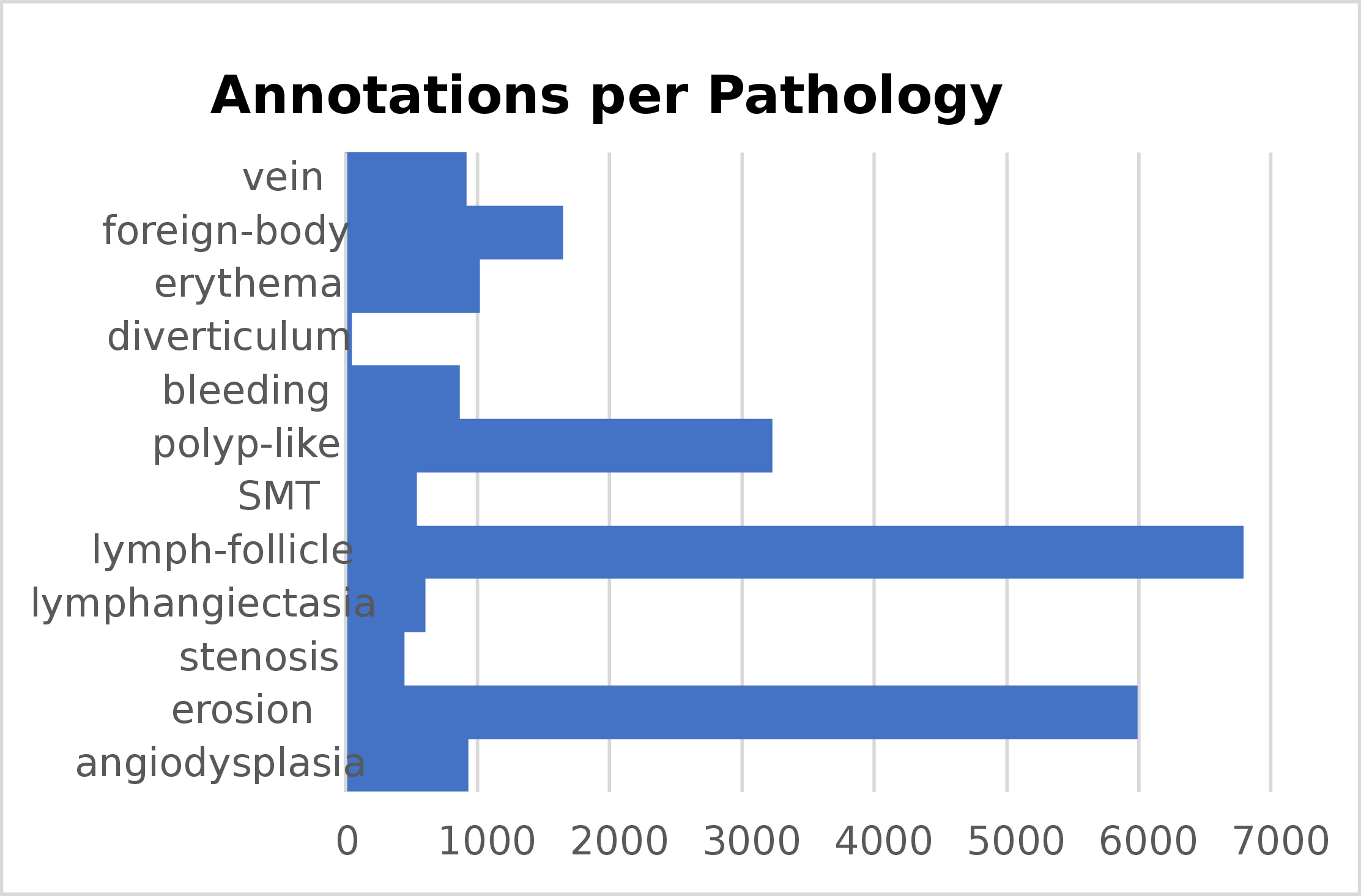}}
  \caption{SEE-AI Dataset Annotations per class. We can observe that classes like lymph-follicle, erosion and polyp-like have more annotations, where as SMT, stenosis, and diverticulum have very few annotations.}
  \label{fig1}
  \vspace*{0.5cm}

\centerline{
\includegraphics[width=0.9\textwidth]
{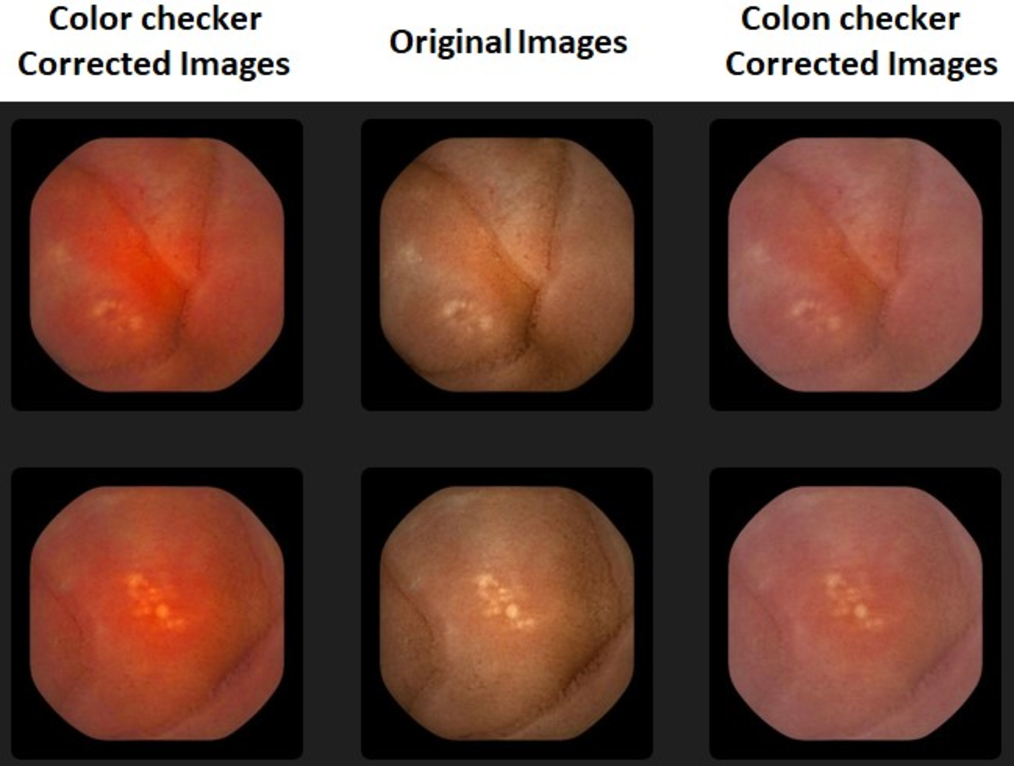}}
  \caption{Original images with their CC and CCC corrected versions. Compared
to the original images, a loss of contrast can be observed with color corrected images.}
  \label{fig2}
\end{figure}

\subsection{Training and performance measures}  
For each case, we train each model using 80\% of data and test with the remaining 20\%. In addition, we ensured that both the training and testing sets have images from all 12 pathology classes. Retinanet and YOLOv5 were trained in each setting  and the best weights were used for the testing.  

The Retinanet method was evaluated using three metrics: F1-score, Intersection over Union (IoU) and Average Precision (AP) with an IoU threshold 0.5. YOLOv5 was evaluated using F1-score and AP. F1-score is a performance metric used for the classification task. It measures the model's accuracy by combining both the precision and recall scores. The precision in Equation \eqref{eq1} is a metric that measures the model's ability to classify a sample as positive.
The recall in Equation \eqref{eq1} is the ability of a classification model to find all the relevant data points. TP is the total number of True Positives for a given class, FP the total number of False Positives, and FN is the total number of False Negatives. The range of F1-score in Equation \eqref{eq3} is between 0 and 1 (1 being the best performance). 

{
\addtolength{\belowdisplayskip}{-0.7ex}
\addtolength{\belowdisplayshortskip}{-0.7ex}
\addtolength{\abovedisplayskip}{-0.7ex}
\addtolength{\abovedisplayshortskip}{-0.7ex}

\begin{align} 
&   \mathit{Precision} = \frac{TP}{TP+FP}  \quad \mathit{Recall} = \frac{TP}{TP+FN}
\label{eq1}
\end{align}

\begin{align} 
&    \mathit{F1} = \frac{2\times\mathit{Precision}\times\mathit{Recall}}{\mathit{Precision}+\mathit{Recall}}
\label{eq3}
\end{align}

\begin{align} 
AP_i = \sum_n (Recall_n - Recall_{n-1})\times Precision_n,
\label{eq4}
\end{align}
\centerline{\text{where $i$ represents the indice for a given image.}}
\centerline{\text{and $n$ represents the thresholds across which TP, FP, TN, FN are determined. }}
The AP for a class $c:= \{x_1, x_2, ..., x_M\}$ is then given by:
\begin{align} 
AP_{c} = \frac{1}M \sum_{i=1}^{M} AP_i.
\label{eq5}
\end{align}
The metrics for each class are weighted based on their distribution within the dataset to compute the overall AP and F1 scores.

\begin{align} 
AP = \sum_{c=1}^{N} AP_c \times w_c ;\quad F1 = \sum_{c=1}^{N} F1_c \times w_c
\label{eq5_1}
\end{align}
 
\begin{align} 
\text{where}, \quad w_c = \frac{M_c}{\sum_{i=1}^{N}M_i}
\label{eq6}
\end{align}

where, $N$ is the total number of classes (pathologies) 
}
\\~\\
 AP is a well-known performance metric which is frequently used to evaluate the performance of the detection model. AP is high when both precision and recall are high, and low when either of them is low across a range of confidence threshold values. The range for AP is between 0 to 1 (1 being the best performance) \cite{AP}. In this work, we report AP with 50\% threshold because 50\% is a reasonable overlap between the predicted and actual bounding boxes and a good indication for a detection model performance \cite{tian2024performance}. F1 and AP in Tables \ref{tab1}-\ref{tab2} are computed using Equation \eqref{eq5}. IoU is the ratio of the overlap area (intersection) to the combined area of prediction and ground truth called \emph{union} \cite{IoU}.
Table~\ref{tab3} presents the performance metrics for IoU across the three experiments and the increase in percentages against the R-OrigD.

\section{Results and Discussion}

The results of the models performance metrics on the test set 
 are summarized in Tables \ref{tab1}- \ref{tab3} and Fig.~\ref{fig3}-~\ref{fig7}. Classes such as \textit{diverticulum and foreign-body} have been exempted from our analysis due to very few and non-diverse samples of the pathologies, that could potentially bias the results.

 \begin{figure}[htbp]
    \begin{minipage}[c]{.48\linewidth}
        \centering       
        \includegraphics[width=1\textwidth] 
        {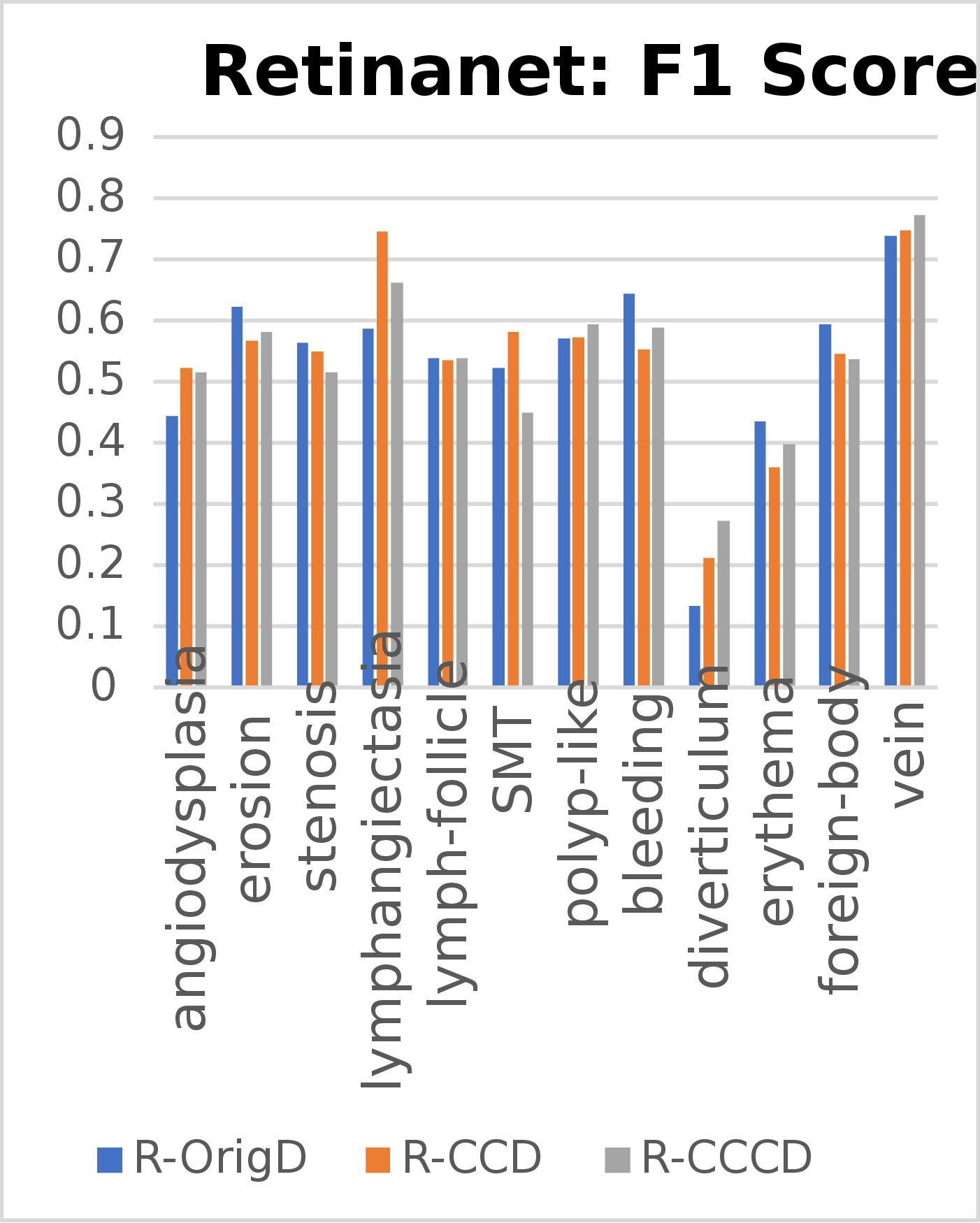}
  \caption{Retinanet F1-scores across different color schemes. The F1 scores increase with both color corrections (CC and CCC) for angiodysplasia, lymphangiectasia and vein. Only CC increases F1-score for SMT while CCC increases F1-score polyp-like. }
  \label{fig3}   
    \end{minipage}
    \hfill%
    \begin{minipage}[c]{.48\linewidth}
        \centering 
        \vspace*{-1\baselineskip}
        \includegraphics[width=1\textwidth] {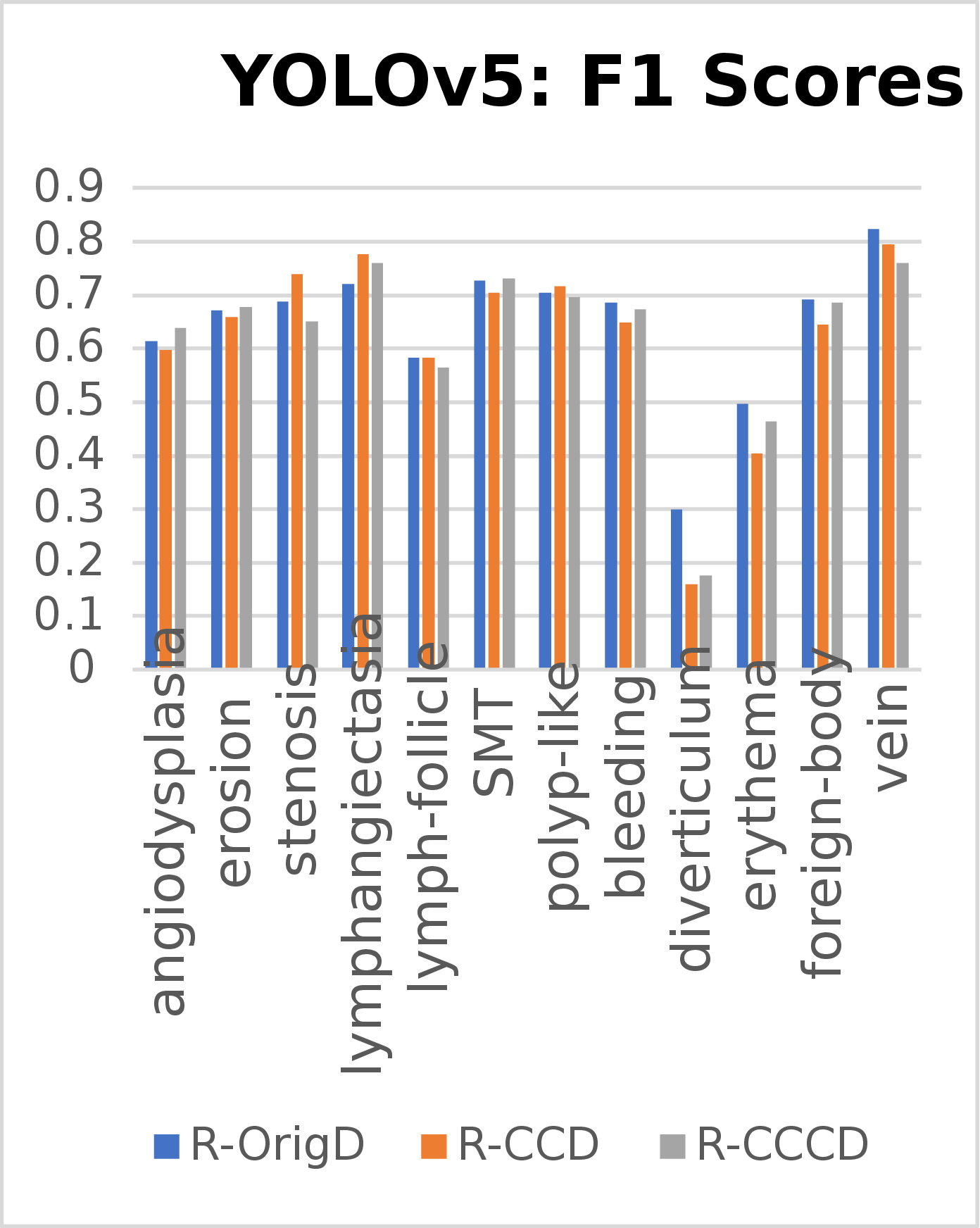}
  \caption{YOLOv5 F1-scores across different  color schemes. The F1 scores increase with both color corrections only for lymphangiectasia. Only CC increases F1-score for stenosis while CCC increases F1-score for angiodysplasia. }
  \label{fig4}   
    \end{minipage}
\end{figure}

\begin{figure}[htbp]
    \begin{minipage}[c]{.48\linewidth}
    \centering
    \vspace*{-1\baselineskip}
       \includegraphics[width=0.9\textwidth]{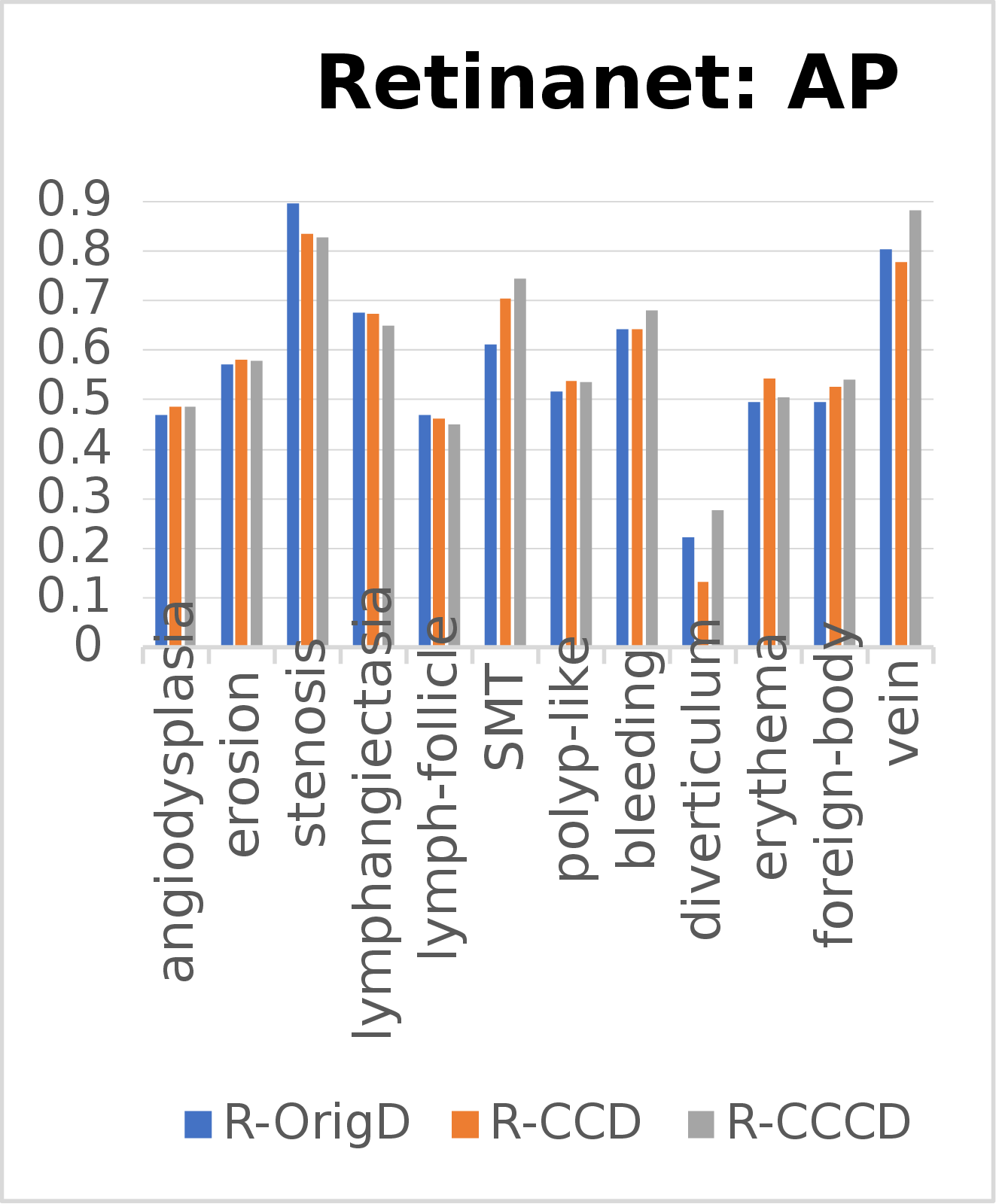}
  \caption{Retinanet AP across different  color schemes. The AP increase with both color corrections (CC and CCC) for angiodysplasia, SMT and polyp-like. Only CC increases AP for Erythema while CCC increases AP for bleeding and vein.}
  \label{fig5}   
    \end{minipage}
    \hfill%
    \begin{minipage}[c]{.48\linewidth}   \centering
       \includegraphics[width=0.9\textwidth]{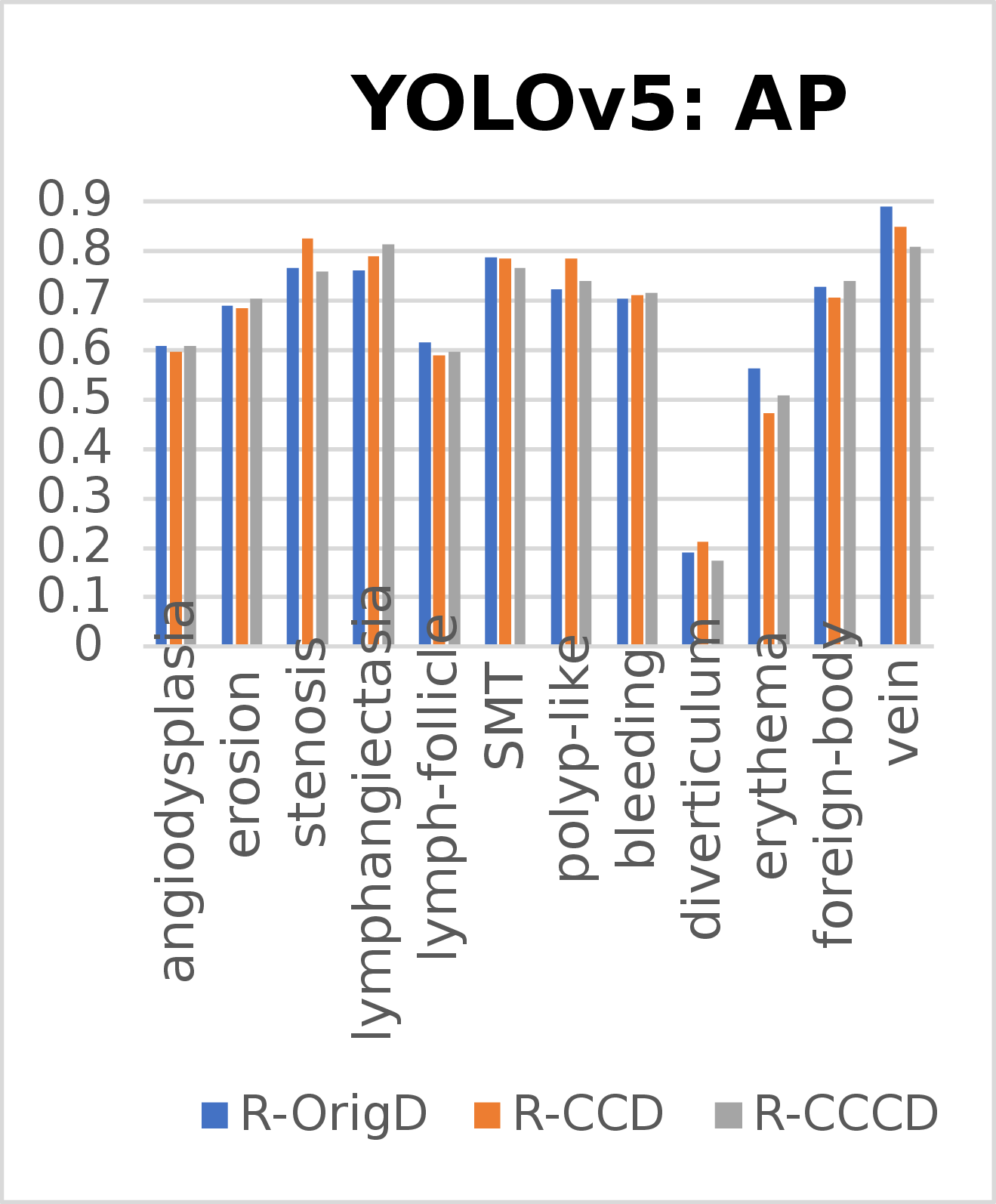}
  \caption{YOLOv5: AP across different color schemes. The AP increase with both color corrections (CC and CCC) for lymphangiectasia, polyp-like and bleeding. Only CC increases AP for stenosis, while CCC increases AP for erosion and angiodysplasia.}
  \label{fig6}   
    \end{minipage}
\end{figure}

\begin{table}[htbp]
\begin{center}
\caption{Models performances in F1-scores across the three different color schemes. Bold values indicate wherever color-correction results in better scores.}
\begin{tabular}{|lc|ccc|ccc|}
\hline
\multicolumn{2}{|l|}{}                                                & \multicolumn{3}{c|}{Retinanet}                                                                   & \multicolumn{3}{c|}{YOLOv5}                                                                      \\ \hline
\multicolumn{1}{|l|}{Pathologies}      & \multicolumn{1}{l|}{Weights} & \multicolumn{1}{l|}{R-OrigD} & \multicolumn{1}{l|}{R-CCD}          & \multicolumn{1}{l|}{R-CCCD} & \multicolumn{1}{l|}{R-OrigD} & \multicolumn{1}{l|}{R-CCD}          & \multicolumn{1}{l|}{R-CCCD} \\ \hline
\multicolumn{1}{|l|}{angiodysplasia}   & 0.036                        & \multicolumn{1}{c|}{0.444}   & \multicolumn{1}{c|}{\textbf{0.523}} & \textbf{0.515}              & \multicolumn{1}{c|}{0.629}   & \multicolumn{1}{c|}{0.601}          & \textbf{0.634}              \\ \hline
\multicolumn{1}{|l|}{erosion}          & 0.252                        & \multicolumn{1}{c|}{0.623}   & \multicolumn{1}{c|}{0.567}          & 0.581                       & \multicolumn{1}{c|}{0.676}   & \multicolumn{1}{c|}{0.667}          & 0.677                       \\ \hline
\multicolumn{1}{|l|}{stenosis}         & 0.021                        & \multicolumn{1}{c|}{0.564}   & \multicolumn{1}{c|}{0.549}          & 0.515                       & \multicolumn{1}{c|}{0.691}   & \multicolumn{1}{c|}{\textbf{0.747}} & 0.657                       \\ \hline
\multicolumn{1}{|l|}{lymphangiectasia} & 0.025                        & \multicolumn{1}{c|}{0.587}   & \multicolumn{1}{c|}{\textbf{0.746}} & \textbf{0.662}              & \multicolumn{1}{c|}{0.718}   & \multicolumn{1}{c|}{\textbf{0.776}} & \textbf{0.759}              \\ \hline
\multicolumn{1}{|l|}{lymph-follicle}   & 0.299                        & \multicolumn{1}{c|}{0.538}   & \multicolumn{1}{c|}{0.535}          & 0.539                       & \multicolumn{1}{c|}{0.584}   & \multicolumn{1}{c|}{0.583}          & 0.563                       \\ \hline
\multicolumn{1}{|l|}{SMT}              & 0.024                        & \multicolumn{1}{c|}{0.522}   & \multicolumn{1}{c|}{\textbf{0.582}} & 0.449                       & \multicolumn{1}{c|}{0.736}   & \multicolumn{1}{c|}{0.705}          & 0.733                       \\ \hline
\multicolumn{1}{|l|}{polyp-like}       & 0.144                        & \multicolumn{1}{c|}{0.570}   & \multicolumn{1}{c|}{0.572}          & \textbf{0.593}              & \multicolumn{1}{c|}{0.697}   & \multicolumn{1}{c|}{\textbf{0.718}} & 0.696                       \\ \hline
\multicolumn{1}{|l|}{bleeding}         & 0.049                        & \multicolumn{1}{c|}{0.643}   & \multicolumn{1}{c|}{0.552}          & 0.589                       & \multicolumn{1}{c|}{0.693}   & \multicolumn{1}{c|}{0.650}          & 0.677                       \\ \hline
\multicolumn{1}{|l|}{diverticulum}     & 0.003                        & \multicolumn{1}{c|}{0.133}   & \multicolumn{1}{c|}{\textbf{0.211}} & \textbf{0.273}              & \multicolumn{1}{c|}{0.301}   & \multicolumn{1}{c|}{0.159}          & 0.175                       \\ \hline
\multicolumn{1}{|l|}{erythema}         & 0.048                        & \multicolumn{1}{c|}{0.435}   & \multicolumn{1}{c|}{0.360}          & 0.397                       & \multicolumn{1}{c|}{0.492}   & \multicolumn{1}{c|}{0.403}          & 0.476                       \\ \hline
\multicolumn{1}{|l|}{foreign-body}     & 0.068                        & \multicolumn{1}{c|}{0.593}   & \multicolumn{1}{c|}{0.545}          & 0.537                       & \multicolumn{1}{c|}{0.697}   & \multicolumn{1}{c|}{0.643}          & 0.680                       \\ \hline
\multicolumn{1}{|l|}{vein}             & 0.031                        & \multicolumn{1}{c|}{0.739}   & \multicolumn{1}{c|}{\textbf{0.747}} & \textbf{0.773}              & \multicolumn{1}{c|}{0.816}   & \multicolumn{1}{c|}{0.797}          & 0.761                       \\ \hline
\multicolumn{1}{|l|}{F1-score}         &                              & \multicolumn{1}{c|}{0.571}   & \multicolumn{1}{c|}{0.553}          & 0.559                       & \multicolumn{1}{c|}{0.649}   & \multicolumn{1}{c|}{0.638}          & 0.638                       \\ \hline
\end{tabular}
\label{tab1}
\end{center}
\end{table}

\begin{table}[htbp]
\begin{center}
\caption{Models performances in AP across the three different experiments schemes. Bold values indicate wherever color-correction results in better scores.}
\begin{tabular}{|lc|ccc|ccc|}
\hline
\multicolumn{2}{|l|}{}                                         & \multicolumn{3}{c|}{RetinaNet}                                                      & \multicolumn{3}{c|}{YOLOv5}                                                         \\ \hline
\multicolumn{1}{|l|}{Pathologies}      & Weights               & \multicolumn{1}{c|}{R-OrigD} & \multicolumn{1}{c|}{R-CCD}          & R-CCCD         & \multicolumn{1}{c|}{R-OrigD} & \multicolumn{1}{c|}{R-CCD}          & R-CCCD         \\ \hline
\multicolumn{1}{|l|}{angiodysplasia}   & 0.036                 & \multicolumn{1}{c|}{0.468}   & \multicolumn{1}{c|}{\textbf{0.486}} & \textbf{0.485} & \multicolumn{1}{c|}{0.627}   & \multicolumn{1}{c|}{0.595}          & 0.608          \\ \hline
\multicolumn{1}{|l|}{erosion}          & 0.252                 & \multicolumn{1}{c|}{0.572}   & \multicolumn{1}{c|}{0.580}          & 0.577          & \multicolumn{1}{c|}{0.695}   & \multicolumn{1}{c|}{0.684}          & \textbf{0.706} \\ \hline
\multicolumn{1}{|l|}{stenosis}         & 0.021                 & \multicolumn{1}{c|}{0.895}   & \multicolumn{1}{c|}{0.834}          & 0.828          & \multicolumn{1}{c|}{0.766}   & \multicolumn{1}{c|}{\textbf{0.827}} & 0.765          \\ \hline
\multicolumn{1}{|l|}{lymphangiectasia} & 0.025                 & \multicolumn{1}{c|}{0.676}   & \multicolumn{1}{c|}{0.672}          & 0.649          & \multicolumn{1}{c|}{0.761}   & \multicolumn{1}{c|}{\textbf{0.790}} & \textbf{0.814} \\ \hline
\multicolumn{1}{|l|}{lymph-follicle}   & 0.299                 & \multicolumn{1}{c|}{0.469}   & \multicolumn{1}{c|}{0.463}          & 0.449          & \multicolumn{1}{c|}{0.622}   & \multicolumn{1}{c|}{0.588}          & 0.597          \\ \hline
\multicolumn{1}{|l|}{SMT}              & 0.024                 & \multicolumn{1}{c|}{0.611}   & \multicolumn{1}{c|}{\textbf{0.703}} & \textbf{0.743} & \multicolumn{1}{c|}{0.791}   & \multicolumn{1}{c|}{0.785}          & 0.765          \\ \hline
\multicolumn{1}{|l|}{polyp-like}       & 0.144                 & \multicolumn{1}{c|}{0.516}   & \multicolumn{1}{c|}{\textbf{0.538}} & \textbf{0.535} & \multicolumn{1}{c|}{0.705}   & \multicolumn{1}{c|}{\textbf{0.756}} & \textbf{0.740} \\ \hline
\multicolumn{1}{|l|}{bleeding}         & 0.049                 & \multicolumn{1}{c|}{0.643}   & \multicolumn{1}{c|}{0.643}          & \textbf{0.679} & \multicolumn{1}{c|}{0.706}   & \multicolumn{1}{c|}{0.709}          & \textbf{0.714} \\ \hline
\multicolumn{1}{|l|}{diverticulum}     & 0.003                 & \multicolumn{1}{c|}{0.222}   & \multicolumn{1}{c|}{0.133}          & \textbf{0.278} & \multicolumn{1}{c|}{0.190}   & \multicolumn{1}{c|}{\textbf{0.213}} & 0.173          \\ \hline
\multicolumn{1}{|l|}{erythema}         & 0.048                 & \multicolumn{1}{c|}{0.496}   & \multicolumn{1}{c|}{\textbf{0.543}} & 0.504          & \multicolumn{1}{c|}{0.536}   & \multicolumn{1}{c|}{0.464}          & 0.497          \\ \hline
\multicolumn{1}{|l|}{foreign-body}     & 0.068                 & \multicolumn{1}{c|}{0.496}   & \multicolumn{1}{c|}{\textbf{0.525}} & \textbf{0.541} & \multicolumn{1}{c|}{0.727}   & \multicolumn{1}{c|}{0.706}          & \textbf{0.739} \\ \hline
\multicolumn{1}{|l|}{vein}             & 0.031                 & \multicolumn{1}{c|}{0.804}   & \multicolumn{1}{c|}{0.778}          & \textbf{0.882} & \multicolumn{1}{c|}{0.889}   & \multicolumn{1}{c|}{0.849}          & 0.809          \\ \hline
\multicolumn{1}{|l|}{AP}               & \multicolumn{1}{l|}{} & \multicolumn{1}{c|}{0.540}   & \multicolumn{1}{c|}{0.548}          & 0.548          & \multicolumn{1}{c|}{0.677}   & \multicolumn{1}{c|}{0.666}          & 0.674          \\ \hline
\end{tabular}
\label{tab2}
\end{center}
\end{table}
\vspace{-1em}

From the analysis of the performance metrics F1-score (Table \ref{tab1}) and AP (Table \ref{tab2}) for both Retinanet and YOLOv5, it is observed that overall, the color corrections have not improved the models detection performance. Consequently, we conduct a detailed investigation into the effect of color correction for specific pathologies such as Erythema, Erosion, Polyp-Like and Bleeding, seeking statistical significance to more precisely attribute the observed effects. To achieve  this, for each model, we analyzed the number of false positives detections, the detected bounding boxes areas (number of pixels), and the percentage of intersection area with the ground truth bounding box. Then, we conducted a combined sign tests for false positives and bounding boxes areas for the selected pathologies between the three different color schemes. The results of p-values of the combined sign test are presented in Tables \ref{tab4}-\ref{tab5}. Green cells indicate where there is a significant difference between different color schemes ( where p-values < 0.05).
We also notice that with both models, there is significant difference between the three color schemes.
All the tables and codes are available on our github link.\\

\begin{table}[htbp]
\begin{center}
\caption{Retinanet: IoU comparison across the three different color schemes. Green arrows indicate wherever there is an increase of the metric value whereas red arrows indicate wherever there is a decrease. }
\begin{tabular}{|l|c|c|c|c|c|}
\hline
Pathologies      & R-OrigD & R-CCD   & R-CCCD  & \% I1  & \% I2 \\ \hline
angiodysplasia   & 0.284   & 0.362  \textcolor{green}{$\uparrow$} & 0.342 \textcolor{green}{$\uparrow$} & 7.80   & 5.80  \\ \hline
erosion          & 0.524   & 0.434 \textcolor{red}{$\downarrow$} & 0.448 \textcolor{red}{$\downarrow$} & -9.00  & -7.60 \\ \hline
stenosis         & 0.367   & 0.338 \textcolor{red}{$\downarrow$} & 0.325 \textcolor{red}{$\downarrow$} & -2.90  & -4.20 \\ \hline
lymphangiectasia & 0.391   & 0.624 \textcolor{green}{$\uparrow$} & 0.487 \textcolor{green}{$\uparrow$} & 23.30  & 9.60  \\ \hline
lymph-follicle   & 0.339   & 0.342 \textcolor{green}{$\uparrow$} & 0.383 \textcolor{green}{$\uparrow$} & 0.30   & 4.40  \\ \hline
SMT              & 0.342   & 0.383 \textcolor{green}{$\uparrow$} & 0.260 \textcolor{red}{$\downarrow$} & 4.10   & -8.20 \\ \hline
polyp-like       & 0.452   & 0.462 \textcolor{green}{$\uparrow$} & 0.486 \textcolor{green}{$\uparrow$} & 1.00   & 3.40  \\ \hline
bleeding         & 0.536   & 0.429 \textcolor{red}{$\downarrow$} & 0.442 \textcolor{red}{$\downarrow$} & -10.70 & -9.40 \\ \hline
diverticulum     & 0.116   & 0.165 \textcolor{green}{$\uparrow$} & 0.200 \textcolor{green}{$\uparrow$} & 4.90   & 8.40  \\ \hline
erythema         & 0.288   & 0.218 \textcolor{red}{$\downarrow$} & 0.246 \textcolor{red}{$\downarrow$} & -7.00  & -4.20 \\ \hline
foreign-body     & 0.468   & 0.365 \textcolor{red}{$\downarrow$} & 0.396 \textcolor{red}{$\downarrow$} & -10.30 & -7.20 \\ \hline
vein             & 0.585   & 0.607 \textcolor{green}{$\uparrow$} & 0.580 \textcolor{red}{$\downarrow$} & 2.20   & -0.50 \\ \hline
\multicolumn{6}{l}{\%I1 is the increase with R-CCD against R-OrigD}
\\ 
\multicolumn{6}{l}{\%I2 is the increase of R-CCCD against R-OrigD}
\end{tabular}
\label{tab3}
\end{center}
\end{table}

 From the p-values, it is observed that there is a significant difference in the number of false positives and bounding boxes areas generated by the models between the three different color schemes. Therefore, color correction significantly impacts the bounding box size and the number of false positives. 
Further analysis of the FP results shows that, on average, R-CCD has significantly more FP compared to R-CCCD and R-OrigD, and that R-CCCD has signficantly more than R-OrigD (Table \ref{tab6}). Analysis for each pathology reveals that the aggregate effect of increased false positives is dependent on the specific characteristics of the pathologies. For Erosion, Erythema, Bleeding and Polyp-like, color-correction schemes generally lead to an increase in the number of false positives generated by the model compared to without color correction (Table \ref{tab6}). For instance, for bleeding, without color correction, there are fewer false positives than after correcting the color. This may be due to the fact that color-corrected images tend to be more saturated, appearing visually more red.  

\begin{table}[htbp]
\centering
 \caption{False Positives: p-values of the combined sign test. In green cells, p-values < 0.05, meaning there  is a significant difference between different color schemes. } 
\begin{tabular}{|l|ccc|ccc|}
\hline
                   & \multicolumn{3}{c|}{Retinanet}                                                                                                                         & \multicolumn{3}{c|}{YOLOv5}                                                                                                                        \\ \cline{2-7} 
\multirow{2}{*}{} & \multicolumn{1}{c|}{R-OrigD}                            & \multicolumn{1}{c|}{R-CCD}                              & R-CCCD                             & \multicolumn{1}{c|}{R-OrigD}                            & \multicolumn{1}{c|}{R-CCD}                              & R-CCCD                         \\ \hline
R-OrigD            & \multicolumn{1}{c|}{X}                                  & \multicolumn{1}{c|}{\cellcolor[HTML]{BFFFBF}7.5267e-19} & \cellcolor[HTML]{BFFFBF}3.8419e-09 & \multicolumn{1}{c|}{X}                                  & \multicolumn{1}{c|}{\cellcolor[HTML]{BFFFBF}1.3781e-04} & \cellcolor[HTML]{BFFFBF}0.0031 \\ \hline
R-CCD              & \multicolumn{1}{c|}{\cellcolor[HTML]{BFFFBF}7.5267e-19} & \multicolumn{1}{c|}{X}                                  & \cellcolor[HTML]{BFFFBF}0.0018     & \multicolumn{1}{c|}{\cellcolor[HTML]{BFFFBF}1.3781e-04} & \multicolumn{1}{c|}{X}                                  & 0.2125                         \\ \hline
R-CCCD             & \multicolumn{1}{c|}{\cellcolor[HTML]{BFFFBF}3.8419e-09} & \multicolumn{1}{c|}{\cellcolor[HTML]{BFFFBF}0.0018}     & X                                  & \multicolumn{1}{c|}{\cellcolor[HTML]{BFFFBF}0.0031}     & \multicolumn{1}{c|}{0.2125}                             & X                              \\ \hline
\end{tabular}
\label{tab4}
\end{table}

\vspace{-0.5cm}
\begin{table}[htbp]
\centering
\caption{Bounding boxes Areas: p-values of the combined sign test.  In green cells, p-values < 0.05 meaning there  is a significant difference between different color schemes.}
\begin{tabular}{|l|ccc|ccc|}
\hline
                   & \multicolumn{3}{c|}{Retinanet}                                                                                                                         & \multicolumn{3}{c|}{YOLOv5}                                                                                                                            \\ \cline{2-7} 
\multirow{2}{*}{} & \multicolumn{1}{c|}{R-OrigD}                             & \multicolumn{1}{c|}{R-CCD}                              & R-CCCD                             & \multicolumn{1}{c|}{R-OrigD}                            & \multicolumn{1}{c|}{R-CCD}                              & R-CCCD                             \\ \hline
R-OrigD            & \multicolumn{1}{c|}{X}                                  & \multicolumn{1}{c|}{\cellcolor[HTML]{BFFFBF}4.1079e-08} & \cellcolor[HTML]{BFFFBF}4.5910e-06 & \multicolumn{1}{c|}{\cellcolor[HTML]{FFFFFF}    X 
   } & \multicolumn{1}{c|}{\cellcolor[HTML]{BFFFBF}0.0094}     & \cellcolor[HTML]{BFFFBF}0.0152     \\ \hline
R-CCD              & \multicolumn{1}{c|}{\cellcolor[HTML]{BFFFBF}4.1079e-08} & \multicolumn{1}{c|}{X}                                  & \cellcolor[HTML]{BFFFBF}0.0142     & \multicolumn{1}{c|}{\cellcolor[HTML]{BFFFBF}0.0094}     & \multicolumn{1}{c|}{\cellcolor[HTML]{FFFFFF}    X 
   } & \cellcolor[HTML]{BFFFBF}0.0056     \\ \hline
R-CCCD             & \multicolumn{1}{c|}{\cellcolor[HTML]{BFFFBF}4.5910e-06} & \multicolumn{1}{c|}{\cellcolor[HTML]{BFFFBF}0.0142}     & X                                  & \multicolumn{1}{c|}{\cellcolor[HTML]{BFFFBF}0.0152}     & \multicolumn{1}{c|}{\cellcolor[HTML]{BFFFBF}0.0056}     & \cellcolor[HTML]{FFFFFF}    X 
    \\ \hline
\end{tabular}

\label{tab5}

\end{table}

\begin{table}[htbp]
\centering
\caption{Total number of False Positives. For the selected pathologies, both
color corrections (R-CCD, R-CCCD)
generally increase the number of false
positives generated by the model compared to without color correction (R-
OrigD). In green cells are the increase in False Positives.}
\begin{tabular}{|l|ccc|ccc|}
\hline
                      & \multicolumn{3}{c|}{Retinanet}                                                                                                                              & \multicolumn{3}{c|}{YOLOv5}                                                                                                                                 \\ \hline
Pathology             & \multicolumn{1}{c|}{R-OrigD} & \multicolumn{1}{c|}{R-CCD}                                              & R-CCCD                                             & \multicolumn{1}{c|}{R-OrigD} & \multicolumn{1}{c|}{R-CCD}                                              & R-CCCD                                             \\ \hline
Erosion               & \multicolumn{1}{c|}{130}     & \multicolumn{1}{c|}{\cellcolor[HTML]{BFFFBF}{\color[HTML]{000000} 250}} & \cellcolor[HTML]{BFFFBF}{\color[HTML]{000000} 225} & \multicolumn{1}{c|}{220}     & \multicolumn{1}{c|}{\cellcolor[HTML]{BFFFBF}{\color[HTML]{000000} 239}} & \cellcolor[HTML]{BFFFBF}{\color[HTML]{000000} 247} \\ \hline
Polyp-like            & \multicolumn{1}{c|}{122}     & \multicolumn{1}{c|}{\cellcolor[HTML]{BFFFBF}{\color[HTML]{000000} 124}} & 90                                                 & \multicolumn{1}{c|}{128}     & \multicolumn{1}{c|}{\cellcolor[HTML]{BFFFBF}{\color[HTML]{000000} 145}} & \cellcolor[HTML]{BFFFBF}{\color[HTML]{000000} 138} \\ \hline
Erythema              & \multicolumn{1}{c|}{148}     & \multicolumn{1}{c|}{\cellcolor[HTML]{BFFFBF}{\color[HTML]{000000} 234}} & \cellcolor[HTML]{BFFFBF}{\color[HTML]{000000} 212} & \multicolumn{1}{c|}{87}      & \multicolumn{1}{c|}{\cellcolor[HTML]{BFFFBF}{\color[HTML]{000000} 93}}  & \cellcolor[HTML]{BFFFBF}{\color[HTML]{000000} 105} \\ \hline
Bleeding              & \multicolumn{1}{c|}{34}      & \multicolumn{1}{c|}{\cellcolor[HTML]{BFFFBF}{\color[HTML]{000000} 65}}  & \cellcolor[HTML]{BFFFBF}{\color[HTML]{000000} 63}  & \multicolumn{1}{c|}{18}      & \multicolumn{1}{c|}{17}                                                 & \cellcolor[HTML]{BFFFBF}{\color[HTML]{000000} 22}  \\ \hline
Total False Positives & \multicolumn{1}{c|}{434}     & \multicolumn{1}{c|}{\cellcolor[HTML]{BFFFBF}{\color[HTML]{000000} 673}} & \cellcolor[HTML]{BFFFBF}{\color[HTML]{000000} 590} & \multicolumn{1}{c|}{453}     & \multicolumn{1}{c|}{\cellcolor[HTML]{BFFFBF}{\color[HTML]{000000} 494}} & \cellcolor[HTML]{BFFFBF}{\color[HTML]{000000} 512} \\ \hline
\end{tabular}
\label{tab6}
\end{table}

\begin{figure}[htbp]
 
    \begin{minipage}[c]{.48\linewidth}
        \centering     
        \includegraphics[width=0.9\textwidth]{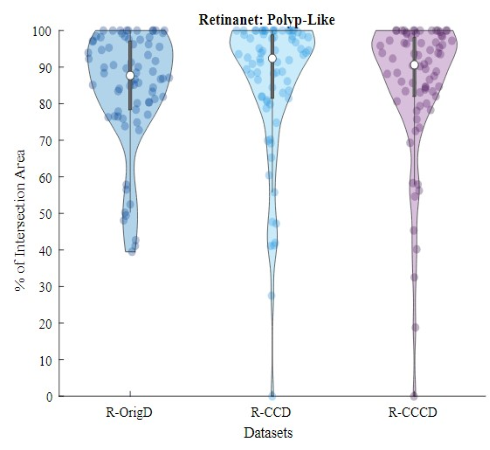}   
    \end{minipage}
    \hfill%
    \begin{minipage}[c]{.48\linewidth}
        \centering     
        \includegraphics[width=0.9\textwidth]{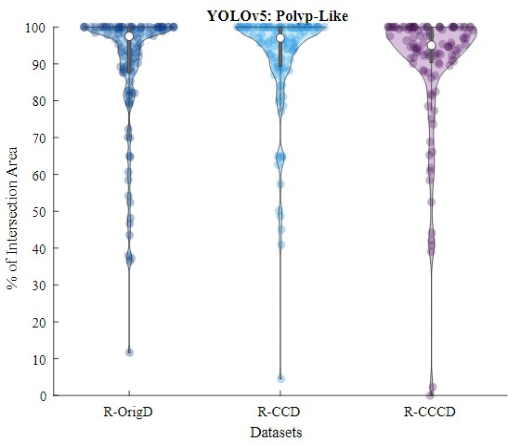} 
    \end{minipage}    
    \caption{Violin plots for Polyp-Like. With Retinanet and YOLOv5, R-CCD creates larger intersection area with the ground truth than the R-CCCD, which creates larger intersection area than the R-OrigD.  } 
     \label{fig8}
 
     \begin{minipage}[c]{.48\linewidth}
        \centering
       \includegraphics[width=0.9\textwidth]{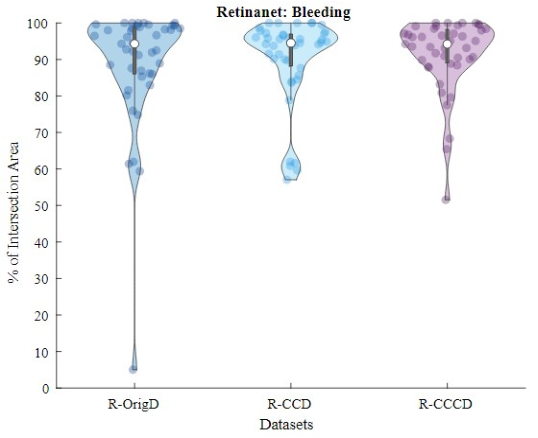}   
    \end{minipage}
    \hfill%
    \begin{minipage}[c]{.48\linewidth}
        \centering
       \includegraphics[width=0.9\textwidth]{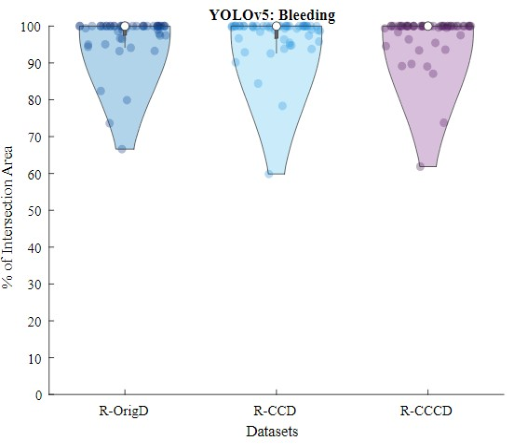}    
    \end{minipage}    
    \caption{Violin plots for Bleeding. With Retinanet and YOLOv5, R-CCCD creates larger intersection area than R-CCD and R-OrigD.}
    \label{fig9} 
\end{figure}

\begin{figure}[htbp]
        \centering
       \includegraphics[width=0.9\textwidth]{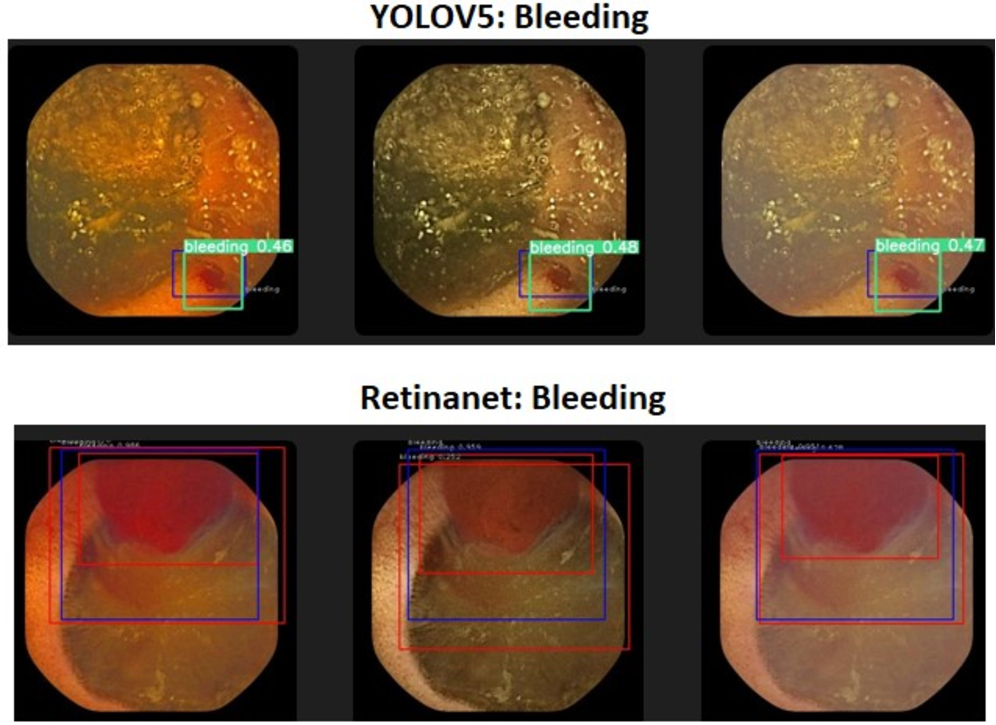}             
       \includegraphics[width=0.9\textwidth]{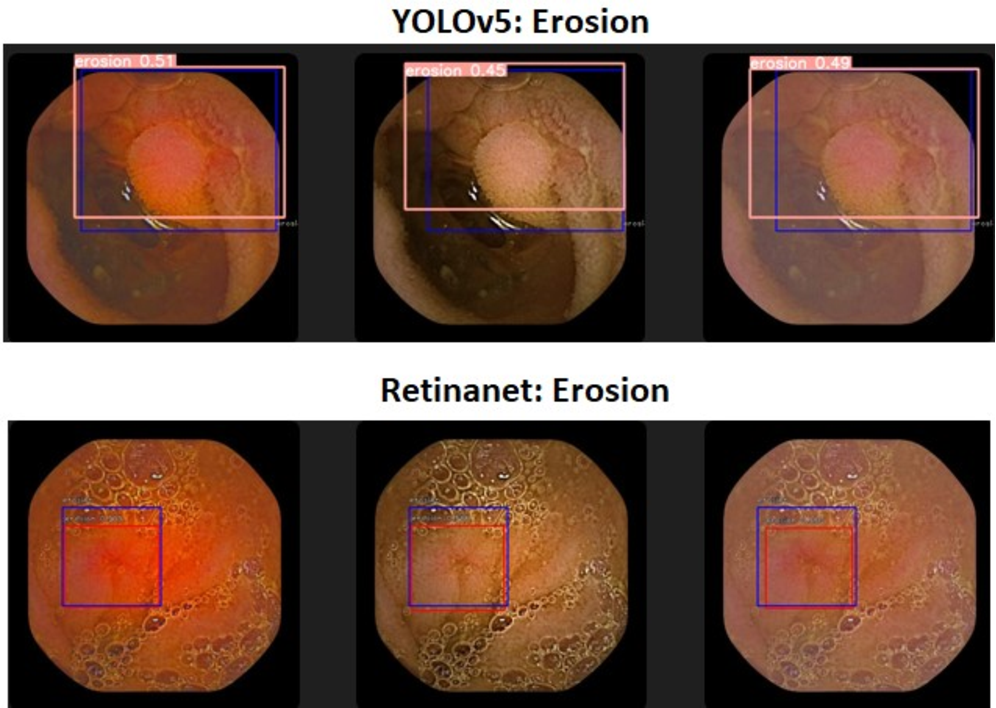}\caption{Models detection results across the three different color schemes. } 
  \label{fig7}
\end{figure}

%
Given that color correction leads models to vary the number of generated bounding boxes compared to the ground truth, with this effect being pathology-dependent and potentially impacting IoU and AP metrics, we analyzed the model's ability to identify all abnormal regions in an image post-color correction. This was done to determine whether the model could still detect abnormal regions regardless of the variation in bounding box count from the original images. To capture the model's response to abnormal regions independently of conventional metrics, we calculated the percentage of area of intersection between the original and generated bounding boxes. We present the violin plots of the percentage of the intersection area of the detected bounding boxes with the ground truth bounding box for different color schemes for Polyp-Like (Figure \ref{fig8}) and Bleeding (Figure \ref{fig9}). From the violin plots, it can be observed the following:
For Polyp-Like, with both Retinanet and YOLOv5, R-CCD creates larger intersection area with the ground truth than the R-CCCD, which creates larger intersection area than the R-OrigD. For Bleeding, with both Retinanet and YOLOv5, R-CCCD creates larger intersection area than R-CCD and R-OrigD. However, with Retinanet, R-CCD creates larger intersection area than R-OrigD where as with YOLOv5, R-CCD creates smaller intersection area than R-OrigD. 
Overall, the color corrections make the models create larger intersection area with the ground truth than the original dataset. 

\section{Conclusion}
This work evaluates the impact of color on the performance of a deep learning model in wireless capsule endoscopy. We use the SEE-AI dataset, create two color corrected sets and benchmark Retinanet and YOLOv5 models on the three datasets. Given the metrics F1 scores, AP and IoU, the impact of color scheme variations on the performance of the object detection model is heterogeneous across different pathologies and color schemes, resulting in no consistent enhancement or deterioration of performance metrics. We further investigated the influence of the color correction on the number of False Positives and the detected bounding boxes. As results, the color corrections generate more false positives and larger intersection areas with the ground truth. However, we observed that the color corrections degrade the contrast of the images in WCE, suggesting that color correction could be most beneficial along with contrast adjustment for diagnostic purposes. This leaves room for future research to investigate more on integrating a color correction scheme with contrast enhancement into deep learning models to evaluate their performance for pathology detection in WCE.

\begin{credits}
\subsubsection{\ackname} 

This work is supported by grants  from ``Quality and Content'' project:324663, ``Capsule'' project:  300031, and ``Capsnetwork'' project: 322600.
\end{credits}

{\small
\bibliographystyle{IEEEtran}
\bibliography{sample}

\begin{thebibliography}{10}
\providecommand{\url}[1]{#1}
\csname url@samestyle\endcsname
\providecommand{\newblock}{\relax}
\providecommand{\bibinfo}[2]{#2}
\providecommand{\BIBentrySTDinterwordspacing}{\spaceskip=0pt\relax}
\providecommand{\BIBentryALTinterwordstretchfactor}{4}
\providecommand{\BIBentryALTinterwordspacing}{\spaceskip=\fontdimen2\font plus
\BIBentryALTinterwordstretchfactor\fontdimen3\font minus \fontdimen4\font\relax}
\providecommand{\BIBforeignlanguage}[2]{{%
\expandafter\ifx\csname l@#1\endcsname\relax
\typeout{** WARNING: IEEEtran.bst: No hyphenation pattern has been}%
\typeout{** loaded for the language `#1'. Using the pattern for}%
\typeout{** the default language instead.}%
\else
\language=\csname l@#1\endcsname
\fi
#2}}
\providecommand{\BIBdecl}{\relax}
\BIBdecl

\bibitem{wang2023global}
R.~Wang, Z.~Li, S.~Liu, and D.~Zhang, ``Global, regional, and national burden of 10 digestive diseases in 204 countries and territories from 1990 to 2019,'' \emph{Frontiers in Public Health}, vol.~11, 2023.

\bibitem{bchir2019multiple}
O.~Bchir, M.~M. Ben~Ismail, and N.~AlZahrani, ``Multiple bleeding detection in wireless capsule endoscopy,'' \emph{Signal, Image and Video Processing}, vol.~13, pp. 121--126, 2019.

\bibitem{neilson2020patient}
L.~J. Neilson, J.~Patterson, C.~von Wagner, P.~Hewitson, L.~M. McGregor, L.~Sharp, and C.~J. Rees, ``Patient experience of gastrointestinal endoscopy: Informing the development of the newcastle endoprem™,'' \emph{Frontline Gastroenterology}, vol.~11, no.~3, pp. 209--217, 2020.

\bibitem{iddan2000wireless}
G.~Iddan, G.~Meron, A.~Glukhovsky, and P.~Swain, ``Wireless capsule endoscopy,'' \emph{Nature}, vol. 405, no. 6785, pp. 417--417, 2000.

\bibitem{muruganantham2021survey}
P.~Muruganantham and S.~M. Balakrishnan, ``A survey on deep learning models for wireless capsule endoscopy image analysis,'' \emph{International Journal of Cognitive Computing in Engineering}, vol.~2, pp. 83--92, 2021.

\bibitem{rahim2020survey}
T.~Rahim, M.~A. Usman, and S.~Y. Shin, ``A survey on contemporary computer-aided tumor, polyp, and ulcer detection methods in wireless capsule endoscopy imaging,'' \emph{Computerized Medical Imaging and Graphics}, vol.~85, p. 101767, 2020.

\bibitem{sushma2022recent}
B.~Sushma and P.~Aparna, ``Recent developments in wireless capsule endoscopy imaging: Compression and summarization techniques,'' \emph{Computers in Biology and Medicine}, vol. 149, p. 106087, 2022.

\bibitem{watine2023enhancement}
L.~Watine, P.~A. Floor, M.~Pedersen, P.~Nussbaum, B.~Ahmad, and {\O}.~Hovde, ``Enhancement of colour reproduction for capsule endoscopy images,'' in \emph{2023 11th European Workshop on Visual Information Processing (EUVIP)}.\hskip 1em plus 0.5em minus 0.4em\relax IEEE, 2023, pp. 1--6.

\bibitem{badano2015consistency}
A.~Badano, C.~Revie, A.~Casertano, W.-C. Cheng, P.~Green, T.~Kimpe, E.~Krupinski, C.~Sisson, S.~Skr{\o}vseth, D.~Treanor \emph{et~al.}, ``Consistency and standardization of color in medical imaging: a consensus report,'' \emph{Journal of digital imaging}, vol.~28, pp. 41--52, 2015.

\bibitem{yokote2024small}
A.~Yokote, J.~Umeno, K.~Kawasaki, S.~Fujioka, Y.~Fuyuno, Y.~Matsuno, Y.~Yoshida, N.~Imazu, S.~Miyazono, T.~Moriyama \emph{et~al.}, ``Small bowel capsule endoscopy examination and open access database with artificial intelligence: The see-artificial intelligence project,'' \emph{DEN open}, vol.~4, no.~1, p. e258, 2024.

\bibitem{liu2021unbiased}
Y.-C. Liu, C.-Y. Ma, Z.~He, C.-W. Kuo, K.~Chen, P.~Zhang, B.~Wu, Z.~Kira, and P.~Vajda, ``Unbiased teacher for semi-supervised object detection,'' \emph{arXiv preprint arXiv:2102.09480}, 2021.

\bibitem{zhou2021instant}
Q.~Zhou, C.~Yu, Z.~Wang, Q.~Qian, and H.~Li, ``Instant-teaching: An end-to-end semi-supervised object detection framework,'' in \emph{Proceedings of the IEEE/CVF Conference on Computer Vision and Pattern Recognition}, 2021, pp. 4081--4090.

\bibitem{mi2022active}
P.~Mi, J.~Lin, Y.~Zhou, Y.~Shen, G.~Luo, X.~Sun, L.~Cao, R.~Fu, Q.~Xu, and R.~Ji, ``Active teacher for semi-supervised object detection,'' in \emph{Proceedings of the IEEE/CVF Conference on Computer Vision and Pattern Recognition}, 2022, pp. 14\,482--14\,491.

\bibitem{redmon2016you}
J.~Redmon, S.~Divvala, R.~Girshick, and A.~Farhadi, ``You only look once: Unified, real-time object detection,'' in \emph{Proceedings of the IEEE conference on computer vision and pattern recognition}, 2016, pp. 779--788.

\bibitem{liu2016ssd}
W.~Liu, D.~Anguelov, D.~Erhan, C.~Szegedy, S.~Reed, C.-Y. Fu, and A.~C. Berg, ``Ssd: Single shot multibox detector,'' in \emph{Computer Vision--ECCV 2016: 14th European Conference, Amsterdam, The Netherlands, October 11--14, 2016, Proceedings, Part I 14}.\hskip 1em plus 0.5em minus 0.4em\relax Springer, 2016, pp. 21--37.

\bibitem{ren2015faster}
S.~Ren, K.~He, R.~Girshick, and J.~Sun, ``Faster r-cnn: Towards real-time object detection with region proposal networks,'' \emph{Advances in neural information processing systems}, vol.~28, 2015.

\bibitem{he2017mask}
K.~He, G.~Gkioxari, P.~Doll{\'a}r, and R.~Girshick, ``Mask r-cnn,'' in \emph{Proceedings of the IEEE international conference on computer vision}, 2017, pp. 2961--2969.

\bibitem{ding2019gastroenterologist}
Z.~Ding, H.~Shi, H.~Zhang, L.~Meng, M.~Fan, C.~Han, K.~Zhang, F.~Ming, X.~Xie, H.~Liu \emph{et~al.}, ``Gastroenterologist-level identification of small-bowel diseases and normal variants by capsule endoscopy using a deep-learning model,'' \emph{Gastroenterology}, vol. 157, no.~4, pp. 1044--1054, 2019.

\bibitem{aoki2020automatic}
T.~Aoki, A.~Yamada, Y.~Kato, H.~Saito, A.~Tsuboi, A.~Nakada, R.~Niikura, M.~Fujishiro, S.~Oka, S.~Ishihara \emph{et~al.}, ``Automatic detection of blood content in capsule endoscopy images based on a deep convolutional neural network,'' \emph{Journal of gastroenterology and hepatology}, vol.~35, no.~7, pp. 1196--1200, 2020.

\bibitem{aoki2021automatic}
------, ``Automatic detection of various abnormalities in capsule endoscopy videos by a deep learning-based system: a multicenter study,'' \emph{Gastrointestinal Endoscopy}, vol.~93, no.~1, pp. 165--173, 2021.

\bibitem{cherepkova2018enhancing}
O.~Cherepkova and J.~Y. Hardeberg, ``Enhancing dermoscopy images to improve melanoma detection,'' in \emph{2018 Colour and Visual Computing Symposium (CVCS)}.\hskip 1em plus 0.5em minus 0.4em\relax IEEE, 2018, pp. 1--6.

\bibitem{lin2017focal}
T.-Y. Lin, P.~Goyal, R.~Girshick, K.~He, and P.~Doll{\'a}r, ``Focal loss for dense object detection,'' in \emph{Proceedings of the IEEE international conference on computer vision}, 2017, pp. 2980--2988.

\bibitem{PillCamSB3}
\BIBentryALTinterwordspacing
M.~logo. Pillcam™ sb 3 capsule endoscopy system. [Online]. Available: \url{https://www.medtronic.com/covidien/en-us/products/capsule-endoscopy/pillcam-sb3-system.html}
\BIBentrySTDinterwordspacing

\bibitem{zhao2020rethinking}
R.~Zhao, B.~Qian, X.~Zhang, Y.~Li, R.~Wei, Y.~Liu, and Y.~Pan, ``Rethinking dice loss for medical image segmentation,'' in \emph{2020 IEEE International Conference on Data Mining (ICDM)}.\hskip 1em plus 0.5em minus 0.4em\relax IEEE, 2020, pp. 851--860.

\bibitem{hsu2020ratio}
W.-Y. Hsu and W.-Y. Lin, ``Ratio-and-scale-aware yolo for pedestrian detection,'' \emph{IEEE transactions on image processing}, vol.~30, pp. 934--947, 2020.

\bibitem{analyticsindiamagAspectRatio}
\BIBentryALTinterwordspacing
S.~Mukherjee, ``{Top ten challenges in object detection every data scientist should know}.'' [Online]. Available: \url{https://analyticsindiamag.com/top-ten-challenges-in-object-detection-every-data-scientist-should-know/}
\BIBentrySTDinterwordspacing

\bibitem{AP}
\BIBentryALTinterwordspacing
A.~Anwar. What is average precision in object detection \& localization algorithms and how to calculate it? [Online]. Available: \url{https://towardsdatascience.com/what-is-average-precision-in-object-detection-localization-algorithms-and-how-to-calculate-it-3f330efe697b}
\BIBentrySTDinterwordspacing

\bibitem{tian2024performance}
J.~Tian, Q.~Jin, Y.~Wang, J.~Yang, S.~Zhang, and D.~Sun, ``Performance analysis of deep learning-based object detection algorithms on coco benchmark: a comparative study,'' \emph{Journal of Engineering and Applied Science}, vol.~71, no.~1, p.~76, 2024.

\bibitem{IoU}
\BIBentryALTinterwordspacing
Kukil. Intersection over union (iou) in object detection \& segmentation. [Online]. Available: \url{https://learnopencv.com/intersection-over-union-iou-in-object-detection-and-segmentation/}
\BIBentrySTDinterwordspacing

\end{thebibliography}
}

%
%
%
%

\end{document}